\pdfoutput=1
\documentclass[titlepage,12pt]{article}

\usepackage{hyperref} 

\usepackage[dvips]{graphicx}

\usepackage[myheadings]{fullpage}
\usepackage{pmetrika}

\usepackage{submit}

\usepackage{setspace}
\usepackage{amsmath, amssymb} 
\usepackage{graphicx, apacite, booktabs}
\usepackage[english]{babel}		
\usepackage{algpseudocode, algorithm, mathrsfs}

\newcommand{\e}[1]{\eqref{#1}}
\newcommand{\bs}[1]{\boldsymbol{#1}}

\DeclareMathOperator*{\argmax}{\arg\!\max}
\usepackage{xcolor}
\usepackage{bm}
\usepackage{multirow}

\usepackage{varioref}
\usepackage{url}

\makeatletter
\AtBeginDocument{%
  \def\doi#1{\url{https://doi.org/#1}}}
\makeatother

\begin{document}
\begin{titlepage}

\title{Multidimensional Item Response Theory in the Style of Collaborative Filtering}
\author{Yoav Bergner$^+$, Peter F Halpin$^*$ \& Jill-J\^enn Vie$^\times$}

\vspace{\fill}\centerline{\today}\vspace{\fill}

\comment{$^+$New York University;  $^*$University of North Carolina, Chapel Hill; $^\times$Inria}

\linespacing{1}

\end{titlepage}

\setcounter{page}{2}

\vspace*{2\baselineskip}

\RepeatTitle{Multidimensional Item Response Theory in the Style of Collaborative Filtering }\vskip3pt

\linespacing{1.5}

\abstracthead
\begin{abstract}
This paper presents a machine learning approach to multidimensional item response theory (MIRT), a class of latent factor models that can be used to model and predict student performance from observed assessment data. Inspired by collaborative filtering, we define a general class of models that includes many MIRT models. We discuss the use of penalized joint maximum likelihood (JML) to estimate individual models and cross-validation to select the best performing model. This model evaluation process can be optimized using batching techniques, such that even sparse large-scale data can be analyzed efficiently. We illustrate our approach with simulated and real data, including an example from a massive open online course (MOOC). The high-dimensional model fit to this large and sparse dataset does not lend itself well to traditional methods of factor interpretation. By analogy to recommender-system applications, we propose an alternative ``validation'' of the factor model, using auxiliary information about the popularity of items consulted during an open-book exam in the course.  

\end{abstract}\vspace{\fill}

\pagebreak

\section{Introduction}

The widespread use of information technology in education has ushered in new research fields such as learning analytics and educational data mining. Within learning analytics, some problems deal explicitly with modeling student ability and growth---longstanding topics of psychometric research---however the methods of estimation have been largely adopted from machine learning and artificial intelligence frameworks \cite{Reye2004,Martin10,Chrysafiadi2013}. Some authors have indeed made direct connections between machine learning approaches and psychometric frameworks like item response theory (IRT), \cite{Desmarais2005,cen2006learning, Pelanek2016}, and the value of established theories of measurement are generally becoming more widely recognized in learning analytics \cite{Bergner2017}. In this paper we describe one particular bridge between these two fields of study by providing an interpretation of multidimensional item response theory (MIRT) that is rooted in machine learning, specifically model-based collaborative filtering \cite{Bergner2012, Hu2009, Su2009, Zhou2008}. The overall goals of this paper are to show how collaborative filtering provides a general framework for predictive MIRT, to discuss the strengths and shortcomings of this approach vis-\`a-vis existing psychometric literature, and to illustrate its application with simulated and real data examples.

We motivate a class of models which contains many well known IRT models (e.g. the 1PL, 2PL, M2PL) as well as a number of models not previously named. The focus in this paper is on compensatory multidimensional models, although this is not an intrinsic limitation of collaborative filtering. The model space can be systematically searched by ordering the candidate models in terms of their parametric complexity. The situation here is operationally similar to exploratory factor analysis (EFA), in which a $r$-dimensional model is nested within all models of dimension greater than $r$. However, unlike conventional approaches to EFA, the dimensionality is determined by out-of-sample prediction accuracy rather than in-sample goodness of fit.  This results in a maximally predictive, but not necessarily explanatory, model.

Prediction accuracy is a widely-used criterion of model fit in machine learning and is an intuitive metric for collaborative filtering. In the present setting the goal is to evaluate each model by how well it predicts missing item responses. This is generically referred to as matrix completion, and in the collaborative filtering setting it is often interpreted in terms of recommending content to users. We argue that traditional views of IRT have also been explicitly concerned with the prediction accuracy, even though it has not been commonly applied to assess model fit in IRT. This view point is nicely summarized in Lord's \citeyear{Lord1980} characterization of the purpose of IRT: ``to describe the items by item parameters and the examinees by examinee parameters in such a way that we can predict probabilistically the response of any examinee to any item, even if similar examinees have never taken similar items before" (p. 11). To facilitate the use of prediction accuracy in IRT settings, we describe some novel results on the upper bound of average prediction accuracy for IRT models, which can be useful when interpreting the quality of model predictions.

For model estimation we use a version of joint maximum likelihood (JML) \cite{Lord1968} in which the item and person parameters are penalized using the $L_2$ norm. Our use of penalized JML is motivated by the literature on regularization, which has its early roots in ridge regression \cite{Alldredge1976}. More recently it has become a staple method in machine learning, where its main use is to avoid overfitting of regression models with a large number of possible predictors \cite{Tibshirani1996, Hastie2009}. Here we take an essentially similar approach in our use of penalized JML for predictive MIRT.

There is a growing literature on the use of penalization and related methods in application to MIRT and related models. Many approaches have used penalization (usually the LASSO or its variants) to impose sparsity on the model parameters. For example penalization of the factor loadings can provide a solution to rotational indeterminacy in exploratory models  \cite<e.g.,>{Hirose2015, Jin2018, sun2016latent, Trendafilov2015, Trendafilov2017}, and penalization of the residual covariances can provide an efficient approach to model modification in confirmatory settings \cite<e.g.>{Pan2017, Pan2019}. In the present research, we do not focus on the issue of factor rotation, but instead on the probabilistic recovery of missing elements of the item response matrix. 

JML is known to suffer from the Neyman-Scott problem \cite{Haberman1977} and has been generally eschewed in the psychometric literature, at least since the marginal maximum likelihood (MML) was made tractable by Bock \& Aitken \citeyear{Bock1981}. MIRT with MML is computationally feasible with a moderate number of dimensions \cite<e.g.,>{Cai2010}; however, it is generally not applicable to the size of problems encountered in collaborative filtering, in which many millions of persons respond to many thousands of items \cite{Zhou2008}. For this reason, collaborative filtering methods have often sacrificed statistical rigor for scalable algorithms and useful solutions, and it is not uncommon to see collaborative filtering implemented via penalized JML but called by other names \cite<e.g.,>{Zhu2016}. Recently Chen, Li, and Zhang  \citeyear{ chen2019joint, Chen2019a} revisited JML in exploratory and confirmatory settings where both the number of persons and the number of items increase without bound. Most relevant to the present study, the authors showed that a constrained version of JML can ensure (a version of) asymptotic consistency of the factor loadings in a model with a fixed number of dimensions.

In this paper, we suggest that penalized JML has a useful role to play in predictive applications that are characteristic of machine learning, and that this role is likely to become more prominent as open educational resources such as massive open online courses  become fixtures of the educational landscape. JML is fast, avoids the necessity of extracting values of the latent variables in a post-processing step, and, when combined with an appropriate penalty function, can balance predictive accuracy with favorability of a sparse factor solution.

In the following section we provide a description of collaborative filtering and consider in more detail its relation to IRT. We then address parameter estimation and the role of regularization, and subsequently address the problem of model selection via out-of-sample prediction accuracy. The data analyses include a data simulation that illustrates model selection and parameter recovery. We also provide two real data applications. The first application is a conventional scale development problem in which we estimate a MIRT model for the Force Concept Inventory \cite{Hestenes1992}. The second application considers data collected from a massive  open online course and calibrates MIRT models up to 20 latent dimensions in a few minutes. We compare our approach to existing computational procedures for JML on both simulated and real datasets. We also obtain convergent evidence for the utility of the learned 20-dimensional model with an automated approach that uses auxiliary data from the MOOC. 


\section{A class of collaborative filtering models for assessment data}

\subsection{Collaborative filtering}

Collaborative filtering is commonly used in so-called recommender systems. Here the goal is to recommend unfamiliar items to a person based on ratings of those items by other persons and prior rating information by the person in question \cite{Su2009}. The Netflix prize, for example, drew much attention to the problem of movie recommendations \cite{Bennett2007}. The basic idea behind collaborative filtering is that when many persons interact with overlapping subsets of items, this information can be used to make inferences about potential new interactions. In practice, we observe a $m \times n$ response matrix $\bm{U} = \{U_{ij}\}$ whose rows represent the response vector of person $i$ to each item $j$. In the usual setting, the ratings are continuous values, for example decimal numbers between 0 and 5. In the case of binary ratings, it is assumed that interactions $U_{ij} \in \{0, 1\}$ for all $i$ and $j$. However, only a small proportion of the possible interactions have been observed, so that $\bm{U}$ is sparse. The problem addressed by collaborative filtering is to predict the missing entries. Therefore, in the usual setting, one can learn a factorization of the response matrix $\bm{U} \simeq \bm{\Theta} \bm{X}^T$ where $\bm{\Theta}$ is a $m \times r$ matrix representing the latent factors of all persons and $\bm{X}$ is a $n \times r$ matrix representing the latent factors of all items. $\bm{\Theta}$ and $\bm{X}$ can be learned from data using alternating least squares or stochastic gradient descent \cite{koren2015advances}, then their product can impute the missing data in $\bm{U}$.

For binary ratings, this problem is solved by choosing a link function. A convenient choice is the logistic function $\sigma$, 
\begin{equation} \label{P}
\sigma(z_{ij}) = \frac{1}{1+e^{-z_{ij}}},
\end{equation}
where the argument $z_{ij}$ is intended to capture the features or covariates of person $i$ and item $j$ that are relevant to predicting the outcome of their interaction. The functional form of $z_{ij}$ and the available covariates depend on the application at hand. 

The basic insight of this paper is that model-based collaborative filtering and IRT have a natural resonance, even if the goals of collaborative filtering and IRT seem fundamentally different. From the perspective of the contemporary psychometrician who uses IRT in scale development, the latent factor structure may be viewed as taking precedence over prediction of unobserved responses, which has only secondary value.  Conversely, the use of collaborative filtering, for example to predict movie reviews, places primary importance on prediction but only secondary importance on the interpretation of factors. While these approaches clearly favor one goal over another, we note that explanatory models are expected to be predictive, while predictive models may also open the door to explanatory insights. As mentioned in the introduction, the role of latent structure and predictive value were seen as entwined in Lord's \citeyear{Lord1980} seminal work on IRT. Contemporary educators tend to believe that mastery of complex subjects requires multiple skills, which may develop at different rates. Predictive models might provide clues to this structure, which can then be refined using standard exploratory and confirmatory methods. We shall illustrate an example of this below using a high dimensional model learned from homework data to predict which items students will reference during an open-book exam.



\subsection{Latent factor models for assessment data}

In collaborative filtering, latent factor models are commonly referred to as singular value decomposition (SVD) models \cite{billsus1998learning}. They are typically encountered in regression form, rather than classification form.

To obtain a probabilistic model of the response matrix $\bm{U}$, let $\bm\theta_i \in \mathbb {R}^r$ be a vector representing features of person $i$ and $\bm{x}_j \in \mathbb {R}^s$ be a vector representing features of item $j$. For each element of $\bm{U}$ define
\begin{equation} \label{cond}
  p_{ij} = \Pr \{U_{ij} = 1 \mid \bm\theta_{i}, \bm{x}_{j}\} = \sigma(z_{ij})
\end{equation}
where $\sigma$ is the logistic function in Equation \e{P} (and $z_{ij}$ is a function of $\bm\theta_{i}$ and $\bm{x}_{j}$). We also require that, conditional on $\bm\theta_{i}$ and $\bm{X}_{j}$, the $U_{ij}$ are i.i.d., yielding the factorization
\begin{equation} \label{ind}
    \Pr \{\bm{U} = {\bf 1} \mid \bm{\Theta}, \bm{X}\} = \prod_{i,j} p_{ij}.
\end{equation}
where $\bm{\Theta} = (\bm\theta_{i})_{1 \leq i \leq m}$, $\bm{X} = (\bm{x}_{j})_{1 \leq j \leq n}$, and {\bf 1} is matrix of ones of the same order as $\bm{U}$. Note that the parametric complexity of both the ``person parameters" and the ``item parameters" grows with the corresponding dimension of $\bm{U}$. 
 
The factorization in Equation \e{ind} can be motivated along the lines of Lord and Novick's \citeyear{Lord1968} discussion of local independence for latent variables. Their basic argument was that the dimension of $\bm\theta_{i}$ could always be chosen to be large enough that local independence holds. This rationale is amenable to predictive applications such as that proposed here, and a same essential rationale underlies the choice of dimensionality in exploratory factor analysis \cite{Bartholomew2011}. 


To simplify this situation, we define the column vectors
\[
    \bm{\theta}_i = \left( 1 \ \ \theta_{i0}  \ \ \theta_{i1} \ \cdots \ \theta_{ir} \, \right)^T  \qquad \bm{x}_j = \left(x_{j0}  \ \ 1  \ \  x_{j1} \ \cdots \ x_{js} \right)^T
\]
and require that $r = s$ so that $\bm{\theta}_i$  and $\bm{X}_j$ are conformable for the inner product
\begin{equation} \label{z}
     z_{ij} = \langle \bm{\theta}_i, \bm{x}_j \rangle = x_{j0} +  \theta_{i0} + \sum_{k=1}^{r} \theta_{i \, k} \; x_{j \, k}.
\end{equation}

Writing the model in its generic form, we have: 

\begin{equation} \Pr\{U_{ij} = 1 \mid \bm\theta_{i}, \bm{x}_{j}\} = \sigma\left(x_{j0} + \theta_{i0} + \sum_{k = 1}^r \theta_{ik} x_{jk}\right)
\end{equation}

This choice of the logit $z_{ij}$, in combination with Equations $\e{P} - \e{ind}$, defines the class of models under consideration in this paper. For each value of $r$, we permit that the item intercept ($x_{j0}$), the person intercept ($\theta_{i0}$), or both may be zero. Specific models are obtained via the choice of $r$ and whether or not to include intercept terms. 
Identification of the sum $z_{ij}$ is discussed in the following section. 

Table~\ref{IRTmodels1} presents a codification of the model space and describes some specific models. For each model, we label the person variables as an ordered pair, where the first component refers to the dimension of $\bm\theta$ and the second indicates whether or not an intercept term is used. Similar notation is used for the item variables.
The logit column describes the form of $z_{ij}$, with person and item subscripts omitted for clarity. The models are presented in slope-intercept form, so that the item intercept is mapped to the usual difficulty parameter for dimension $k$ as $\beta_j = -x_{j0}/x_{j1}$ \cite{Reckase2009}. 
 
\begin{table}[htp]
  \caption{Examples of models with their number of parameters}
  \centering
\begin{tabular}{cccl}
\hline
\# $\bm\theta$ param. & \# $\bm{x}$ param.  & logit & model name \\
\hline
0 & 1 & $x_0$ & person ind. model  \\ 
1 & 0 & $\theta_0$ & item ind. model \\
1 & 1 & $x_0 + \theta_0$ & Rasch (1PL)  \\
1 & 2 & $x_0 + \theta_1 x_1$  &  2PL  \\
2 & 1 & $ \theta_0 + \theta_1 x_1 $  &  non-standard  \\
$r$ & $r+1$ & $x_0 + \sum_k \theta_k x_{k} $  & M2PL\\
$r + 1$ & $r$ & $\theta_0 + \sum_k \theta_{k} x_{k} $  & non-standard\\

\hline
\end{tabular}
\label{IRTmodels1} \\ 

\vspace{5pt}
\raggedright
\emph{Note} The abbreviations 1PL and 2PL refer to one- and two-parameter logistic models, respectively and M2PL the multidimensional 2PL.  
\end{table}%

The correspondence between logit and some IRT models is also indicated. The person independence model was described by \citeA{Holland1990}, which ensures that the proportion of correct responses for each item is fitted perfectly, but does not depend on $\bm\theta$. Its counterpoint is a previously unnamed model, which we refer to the item independence model. It ensures that the proportion of correct responses for each person is fitted perfectly, but does not depend on $\bm{x}$. Both models are uninteresting, in the sense that they do not model dependence beyond first order margins. The Rasch model is obtained by combining the two independence models. The two-parameter logistic (2PL) and Reckase's Multidimensional 2PL (M2PL) are also noted, with their unnamed counterpoints. 

The model space described in Table~\ref{IRTmodels1} has nesting structure analogous to that of exploratory factor analysis. For example both independence models are nested with the Rasch model. The Rasch model is nested within the 2PL also the non-standard counterpoint of the 2PL. For multidimensional models, refering to $r$ as the order of the model, it is apparent that all models of order $r$ are nested within all models of order $r+1$ or higher. Because the  model space is nested, the order of a model can also be used as a measure of its parametric complexity. 

We have not presented a substantive rationale for any of the models defined via Equation \e{z}. From a machine learning perspective, the goal is simply to select the model which has the best out-of-sample prediction of item responses. The model selection problem is then separated into two steps: training and validation. Training refers to estimation of model parameters, and is discussed in the next section. Validation is used both to select the regularization tuning parameters for each model, and also to select the best value of $r$; validation is discussed in the subsequent section. 

\subsection{Parameter estimation and regularization}

Let $\bm{\zeta}$ be a random vector containing the $\bm{\theta}_{i}$ and  $\bm{X}_{j}$. The length of  $\bm\zeta$ is bounded by $(m + n) \times (r + 1)$, and it depends on whether an intercept term is present for the items or the people or both. We also let $f(\bm\zeta)$ denote the probability density function of $\bm\zeta$ and $p_{ij}(\bm\zeta) = \Pr \{U_{ij} = 1 \mid \bm\zeta\}$. Then the model-implied distribution of the observed response matrix $\bm{U}$ can be written 

\begin{equation} \label{MML}
 L(\bm{U}) = \int \prod_{i, \, j}  p_{ij}(\bm\zeta)^{\, U_{ij}} (1-p_{ij}(\bm\zeta))^{\,(1-U_{ij})} f(\bm\zeta) \, d \bm\zeta.
\end{equation}
The integral in Equation \e{MML} is intractable, so that any estimation strategy requiring the marginalization over the model parameters (or a subset thereof) involves numerical integration. Cai \citeyear{Cai2010} reviews available methods when the item parameters are treated as fixed (i.e., non-random). The case where the parameters of both items and persons are treated as random is discussed from a Bayesian perspective by \citeA{cho2011alternating}. These approaches are not computationally attractive when searching over a large model space, and therefore we seek a compromise. 

Conditional ML has traditionally been used to avoid computational challenges such as those in Equation \e{MML}. In general, conditional ML refers to any maximum likelihood estimator in which a conditional density of the data is used in lieu of its marginal density \cite{Palmer2004}. In IRT literature, the title ``conditional ML" has been reserved for the method proposed by \citeA{Andersen1970}, while estimation based on Equation \e{cond} is referred to as JML. Despite the lack of popularity of JML, as discussed in the introduction, we advocate its use in the current application because it is a computationally more efficient approach to the problem at hand. 

The log of the JML function is 
\begin{equation}
\label{logl}
\ell (\bm\zeta; \bm{U}) =   \sum_{i, j}\left( U_{ij} \log p_{ij}(\bm\zeta) + (1-U_{ij})\log(1-p_{ij}(\bm\zeta)) \right).
\end{equation}
In contrast to marginal approaches, $\bm\zeta$ is treated as a parameter vector to be estimated rather than a random variable. There remain difficulties in application of this approach. First, since $\bm\zeta$ is a bilinear function of the model parameters, $\ell$ is not concave and this complicates numerical optimization of Equation \e{logl}. Marginal approaches avoid this problem, because they require numerical optimization only over the items parameters. Block-wise coordinate descent methods similar to that originally proposed by \citeA{Birnbaum1968} have been developed for bilinear logistic regression, but these only guarantee convergence to local maxima \cite{Shi2014}. Cai \citeyear{Cai2010} suggested addressing this problem of multiple maxima by choosing starting values for item parameters based on an initial EFA of the item tetrachoric correlation matrix. This initialization can be applied to the present approach as well, but random initialization appears to perform equally well. In terms of time to convergence, one trades slightly longer time to optimize parameters for time saved by skipping the EFA stage.

A second difficulty has to do with rotational indeterminacy. It is well known that $z_{ij}$ is subject to rotation by a $k$-by-$k$ invertible matrix $\bm T$ \cite<e.g.,>{Browne2001, Reckase2009}. There are therefore a total of $k^2$ constraints in the $k$-dimensional case. It is usual to impose $k(k+1)/2$ constraints by requiring that ${\rm COV} \{\bs{\theta}\} = \bm{I}$, which implies orthogonality, $\bm T^{-1} = \bm T'$. The remaining $k(k-1)/2$ constraints are imposed on the $\bm{X}_j$. Typically, an interpretable choice is pursued once the model has been estimated.

One way that the problem of model complexity has been addressed in the machine learning literature is under the rubric of sparsity (for models, not data) \cite{Lan2013, Shi2014, Sahebi2016, Sahebi2019}. The idea is to penalize non-zero model parameters using standard approaches developed in regression \cite{Tibshirani1996, Fan2001}. In the present application, we subtract a penalty term from Equation \e{logl}. Our objective function to maximize becomes:
\begin{equation}
\label{objective}
\ell (\bm\zeta; \bm{U}) =   \sum_{i, j}\left( U_{ij} \log p_{ij}(\bm\zeta) + (1-U_{ij})\log(1-p_{ij}(\bm\zeta)) \right) - \lambda \left( \sum_i || \bm{\theta}_i ||^2_2 + \sum_j || \bm{x}_j ||^2_2 \right)
\end{equation}
where $|| \cdot ||^2_2$ denotes the squared $L^2$ norm and  $\lambda$ is a hyper-parameter to be determined by cross-validation. Taking this approach, every non-zero model parameter (other than the person and item intercepts) contributes to the penalized likelihood function. However, this does not guarantee that the resulting parameterization is free from rotational indeterminacy. In particular, it is obvious that a change of sign will not affect the penalty terms. Penalties involving crossproducts, for example those used for factor rotation in EFA \cite{Browne2001}, are not concave and, in our experience these complicate numerical optimization. Our objective is similar to the constrained JML estimation in \cite{chen2019joint}. The authors there chose to constrain the optimization in a compact $\forall i, j, || \bm{\theta}_i ||^2 \leq C^2, || \bm{X}_j ||^2 \leq C^2$ where $C = 5\sqrt{r}$ is pre-specified but can be tuned. Because we use a penalized likelihood rather a constrained parameter space, we do not need to make a projection on a feasible set of parameters at each step, and $\lambda$ is selected using cross-validation.



\section{Metrics and cross-validation}
\label{CVpred}

Two issues arise in selecting the best model from the point of view of prediction accuracy: non-convexity of the likelihood surface and overfitting of a too-complex model to the data. Multiple restarts are usually used to address the convexity problem, and tuning of the penalization hyper-parameters addresses the overfitting problem. We now describe two algorithms for carrying out such a process. We first give an overview of the two approaches and then describe details below.

The first approach involves a traditional division of the dataset row-wise into a training set, a cross-validation set, and a test set. We refer to it as the striated method. Recall that our prediction problem differs from a simple classification or regression problem where a vector of features is used to generate a prediction. In our case, it is a matrix completion problem where some entries $U_{ij}$ are known and other entries $U_{ij}$ should be inferred. As we are splitting the dataset row-wise, it is different from the usual setting in collaborative filtering. Once a model is learned, i.e. the hyper-parameters and item parameters estimated, a hold-out procedure (e.g. $n$-fold cross-validation) must still be carried out as a performance test. Some items are held out, person parameter estimates are derived based on the remaining items, predictions are made about the held-out items, and finally the performance on those items is checked. The process is repeated for different sets of held-out items, $n$ times if $n$ folds are used.

In the striated algorithm, this $n$-fold performance test is carried out twice. First, cross-validation is applied on a validation set for finding the best hyper-parameters. Then, cross-validation is applied on the test set. Models are refit using training, validation data, and test folds until convergence of the penalized joint maximum likelihood. Finally, performance is evaluated using area under the ROC curve (AUC) \cite{bradley1997use} as described in the algorithm details below.

Given the person-item interaction nature of the prediction problem, another approach to learning the hyper-parameters and item parameters is to sample elementwise rather than row-wise for holding out data. 
Training and cross-validation happen at the same time, which may be seen as a disadvantage in that the effects of local maxima and choice of hyper-parameters are conflated. But from the point of view of AUC, this is not necessarily a concern. In the elementwise method, during each restart, a random sample of matrix elements are held out (treated as missing). Since both person and item parameters are estimated, the performance on the held-out items may be evaluated for each restart. After a preset number of restarts have been performed, the prediction performance is averaged (as in the case of $n$ fold cross-validation). Finally these values are used on a test set of new persons, as before.

For both methods, in the first stage, a validation set is available. Therefore the optimization of the penalized joint maximum likelihood can be stopped whenever the performance score on the validation set is not increasing anymore; such a technique has been referred to as early stopping \cite{prechelt1998early,yao2007early}.

\subsection{Formalization of the algorithms}

Let the space of models be denoted by $\Omega = (\omega_j)_j$, and the space of hyper-parameters under consideration for each model $\omega_j$ as $\Lambda_j = (\lambda_{ij})_{1 \leq i \leq n_j}$. Let there be an estimation procedure ESTIM (e.g. JMLE) which, given a response matrix $\bm{\bm{U}}$, a model $\omega$, and a vector of hyper-parameters $\bm{\lambda}$, returns a vector of estimated item parameters $\bm{X}$ and person parameters $\bm\Theta$. Functionally, $$(\bm{X}, \bm\Theta) \gets \textrm{ESTIM}(\bm{U},\omega,\bm{\lambda}).$$

Once the parameters have been estimated, we can carry out a performance test PT that computes a score $S$ (usually area under the ROC curve) between the predicted probabilities for the non-missing elements of $\bm{U}$ and the actual elements. $$S \gets \textrm{PT}(\bm{U},\bm{X},\bm\Theta,\omega,\bm{\lambda}).$$

Let there be a striated performance test SPT, which given a response matrix $\bm{U}$, a model $\omega$, a vector of hyper-parameters $\bm{\lambda}$, returns a prediction performance measure $S$ by means of $k$-fold cross-validation on the items of $\bm{U}$. That is, the items (columns) of $\bm{U}$ are divided into $k$ equal-sized sets $\bm{U}_{val}^{(1)}, \ldots, \bm{U}_{val}^{(k)}$ (``folds"); abilities  for persons (rows of $\bm{U}$) are estimated by maximizing the likelihood and excluding the items in one fold at a time. Probabilities for correct response are estimated on the items in the held-out fold, and the performance error is then computed. Finally, the performance error is averaged over all folds, see Algorithm~\ref{striated}. For now, the number of cross-validation folds $k$ is taken to be a global parameter, rather than a variable argument in SPT. Thus, $$S \gets \textrm{SPT}(\bm{U}_{tr}, \bm{U}_{cv}, \omega, \bm{\lambda}).$$


\subsubsection*{Striated method}

One algorithm for tuning the hyper-parameters within a model $\omega_j$ and selecting the best model within the space $\Omega$ is described in pseudocode in Algorithm~\ref{striated}. For this algorithm, it is assumed that one has divided the $\bm{U}$ matrix row-wise into a training set $\bm{U}_{tr}$, cross-validation set, $\bm{U}_{cv}$, and test set, $\bm{U}_{ts}$ (proportions may be, for example, 50\%/25\%/25\%).


\begin{algorithm}[ht]
  \hspace*{\algorithmicindent} \textbf{Input:} $\bm{U}_{train}, \bm{U}_{val}$, model $\omega$, $\bm{\lambda}$ regularization hyper-parameter\\
  \hspace*{\algorithmicindent} \textbf{Output:} validation score $S$
  \begin{algorithmic}[1]
    \Procedure{SPT}{$\bm{U}_{tr}, \bm{U}_{cv}, \omega, \bm{\lambda}$}
      \For{$n$ = 1 to $N_f$} \Comment{for each fold}
        \State $\bm{X}^{(n)}, \bm\Theta^{(n)} \gets $ \textrm{ESTIM}($\bm{U}_{tr} \cup \bm{U}_{cv} \setminus U_{fold}^{(n)}, \omega, \bm{\lambda})$
        \State $S^{(n)} \gets$ \textrm{PT}($U_{fold}^{(n)}, \bm{X}^{(n)}, \bm\Theta^{(n)}, \omega, \bm{\lambda}$)
      \EndFor
      \State $S \gets 1/N_f \sum_n S^{(n)}$  \Comment{average score over folds}
      \State \Return $S$
    \EndProcedure\bigskip
  \end{algorithmic}

  \hspace*{\algorithmicindent} \textbf{Input:} candidate models $\Omega$, hyper-parameters $\Lambda_j$ for each model $\omega_j$\\
  \hspace*{\algorithmicindent} \textbf{Output:} best model $\omega^*$
  \begin{algorithmic}[1]
        \Procedure{BestModelStriated}{$\Omega, (\Lambda_j)_j$}
      \For{$\omega_j$ in $\Omega$}
        \For{$\lambda_{ij}$ in $\Lambda_j$}
          \State $S_{cv}^{(i)} \gets$ \textrm{SPT}($\bm{U}_{tr}, \bm{U}_{cv}, \omega_j, \lambda_{ij}$)
        \EndFor
        \State $\lambda^{*}_j \gets  \lambda_{ij}, \quad i = \argmax S_{cv}^{(i)}$ \Comment{best hyper-parameter $\lambda$ given $\omega_j$}
            \State $S_{test}^{(j)} \gets$ \textrm{SPT}($\bm{U}_{tr} \cup \bm{U}_{cv}, \bm{U}_{ts}, \omega_j, \lambda^{*}_{j}$)
      \EndFor
      \State $\omega^{*} \gets \omega_j, j =  \argmax S_{test}^{(j)}$ \Comment{best model, $\omega^*$}
      \State \Return $\omega^*$
    \EndProcedure
  \end{algorithmic}
 \caption{Striated method}\label{striated}
\end{algorithm}

\subsubsection*{Elementwise method}

In Algorithm~\ref{striated}, there is a linear progression between estimating item parameters, tuning the hyper-parameters on a cross-validation set, and finally selecting the best model. Each of these steps requires a unique portion of the data $\bm{U}$, the training set $\bm{U}_{tr}$, the cross-validation set $\bm{U}_{cv}$, and the test set $\bm{U}_{ts}$. 

In a variation of this algorithm, the first two steps are rearranged in such a way that only one division of $\bm{U}$ into a training set $\bm{U}_{tr}$ and test set $\bm{U}_{ts}$ is required. Fits to the training set $\bm{U}_{tr}$ are actually based on random elementwise samples from the training set. The unsampled data constitute a cross-validation set, which changes for each random restart. The average performance over restarts is used to determine the best hyper-parameter $\lambda_j^*$ first, for a given model. The final step is as before, where parameters are re-estimated using all of $\bm{U}_{tr}$ and fixed $\lambda_j^*$ along with the striated performance test on the test set.

We explain the elementwise procedure in more detail, beginning with definitions and culminating in Algorithm~\ref{elementwise}. 

Let the dimensions of $\bm{U}$ be $n \times m$. Define a random matrix $B$ with the same dimensions, and for a given probability $p$, such that

\begin{equation}
 B_{ij}(p) = \begin{cases} 
    1 & \mbox{with probability} \quad p \\
      0 & \mbox{with probability} \quad  1-p.
   \end{cases}
\end{equation}


$B$ may be used as an indicator matrix for sampling from $\bm{U}$ each time ESTIM is restarted. We define the sample matrix $\bm{U}^{(s)}$ and its complement $\bm{U}^{(c)}$\begin{equation}%
\begin{minipage}{0.35\linewidth}%
\[ \bm{U}^{(s)}_{ij} = \begin{cases} 
    U_{ij} & \mbox{if } B_{ij} =1 \\
      \mbox{NA} & \mbox{otherwise} 
   \end{cases}
\]%
\end{minipage}
\begin{minipage}{0.35\linewidth}
\[ \bm{U}^{(c)}_{ij} = \begin{cases} 
    U_{ij} & \mbox{if } B_{ij} =0 \\
      \mbox{NA} & \mbox{otherwise.} 
   \end{cases}
\]%
\end{minipage}%
\end{equation}%
$\bm{U}^{(s)}$ has roughly proportion $p$ non-missing matrix elements (e.g. $p=0.7$). The function ESTIM regards the missing matrix elements in $\bm{U}^{(s)}$ as missing at random. The log-likelihood and parameter estimates are obtained as usual. $$(\bm{X}, \bm\Theta) \gets \textrm{ESTIM}(\bm{U},\omega,\bm{\lambda}).$$

We test the performance of the parameters estimated via $\bm{U}^{(s)}$ on the complementary matrix elements $\bm{U}^{(c)}$.

\begin{algorithm}[ht]
  \hspace*{\algorithmicindent} \textbf{Input:} candidate models $\Omega$, hyper-parameters $\Lambda_j$ for each model $\omega_j$\\
  \hspace*{\algorithmicindent} \textbf{Output:} best model $\omega^*$
  \begin{algorithmic}[1]
    \Procedure{BestModelElementwise}{$\Omega, (\Lambda_j)_j$}
      \For{$\omega_j$ in $\Omega$}
        \For{$\lambda_{ij}$ in $\Lambda_j$}
          \For{$n$ = 1 to $N_r$} \Comment{for each restart}
            \State $B^{(n)} \gets $  random matrix $B(p)$
            \State $\bm{X}^{(n)}, \bm\Theta^{(n)} \gets $ \textrm{ESTIM}($\bm{U}^{(s,n)}_{tr}, \omega_j, \lambda_{ij})$
            \State $S_{ij}^{(n)} \gets$ \textrm{PT}($\bm{U}^{(c,n)}_{tr}, \bm{X}^{(n)}, \bm\Theta^{(n)}, \omega_j, \lambda_{ij}$)
          \EndFor
        \EndFor
        \State $S_{ij} \gets 1/N_r \sum_n S_{ij}^{(n)}$  \Comment{average score over elementwise samples}
        \State $\lambda^{*}_j \gets  \lambda_{ij}, \quad i = \argmax S_{ij} $ \Comment{best hyper-parameter $\lambda$, given $\omega_j$}
        \State $S_{test}^{(j)} \gets$ \textrm{SPT}($\bm{U}_{tr}, U_{test}, \omega_j, \lambda^{*}_{j}$)
      \EndFor
      \State $\omega^{*} \gets \omega_j, j =  \argmax S_{test}^{(j)}$ \Comment{best model, $\omega^*$}
      \State \Return $\omega^*$
    \EndProcedure
  \end{algorithmic}
 \caption{Elementwise method}\label{elementwise}
\end{algorithm}

\subsection{Prediction accuracy}
\label{upper-bound}

Before proceeding to our applications, we wish to say a few words about using prediction accuracy as a performance measure. Prediction accuracy, i.e., the proportion of person-item elements correctly classified as 0 or 1, is a simple and intuitive measure, but it comes with limitations. Operationally, given the model and estimated parameters, each matrix element is understood to be a random draw with a certain expectation value. If the probability for the matrix element is estimated at or above 0.5, then a 1 is predicted. Prediction accuracy is not sensitive to degree of confidence, as either a cross-entropy loss (equivalent to negative log-likelihood) or a root mean squared error would be. For example, predicting a correct response with a probability of 0.51 or a probability of 0.95 is penalized the same if the response was observed incorrect. 

Importantly, even if ``true'' parameter values were known, prediction accuracy would not of course be close to unity, unless each item was either infinitely too easy or too hard for each person. There is an upper limit to the prediction accuracy that can be expected, which helps to put into perspective whether incremental increases in performance using high-dimensional models should be regarded as significant improvements. This limit is not accessible ahead of time (except in a simulation study) as it depends both on the model and on the distribution of item and person parameters. As we are not aware of prior analytic results on the expected attainable accuracy for item response models, we have included a derivation and example below.

If the distribution of random variable $P$ in the expectation matrix is given by a distribution function  $f_P(p)$, then the expected accuracy score is given by the following ``average''

\begin{equation}
\label{sbest}
S = \int_0^{0.5} (1-p) f_P(p) dp + \int_{0.5}^{1} p f_P(p) dp 
\end{equation}

\noindent where the first term accounts for predicted-to-be-wrong and the second term for predicted-to-be-right matrix elements. As stated, the shape of $f_P(p)$ in turn depends on the distribution of the person and item parameters and the link function in the model. As an explicit  example, for the Rasch or 1PL model, the probability of a correct response is distributed according to Equation \e{Raschdistro}.

\proposition{Let $\Theta$ and $\Delta$ be independent random variables with densities $f_{\Theta}(\theta)$ and $f_{\Delta}(\delta)$. Given the model,
\begin{equation}
\label{Rasch}
P = g(\Theta,\Delta) = \frac{1}{1+e^{-(\Theta-\Delta)}},
\end{equation}
then
\begin{equation}
\label{Raschdistro}
f_P(p) = \frac{1}{p(1-p)} \int_{-\infty}^{\infty} f_{\Theta}(\theta) f_{\Delta}\left(\theta+\ln\frac{1-p}{p}\right) d\theta.
\end{equation}
}

\proof{By definition, the density
\begin{equation}
f_P(p) = \frac{d}{dp} \Pr[P  \le p] =  \frac{d}{dp} \iint_{D_p} f_{\Theta}(\theta)  f_{\Delta}(\delta) d\theta d\delta,
\end{equation}
where $D_p = \{(\theta,\delta) \mid g(\theta,\delta) \le p\}$ is the region of integration in the $(\theta,\delta)$-plane.
Inverting Equation \e{Rasch}, we can write this constraint as
\begin{equation}
\delta \ge \theta + \ln \frac{1-p}{p}.
\end{equation}
Absorbing the constraint into the integral limits,
\begin{eqnarray}
f(p) &=&  \frac{d}{dp} \int_{-\infty}^{\infty} \int_{\theta + \ln \frac{1-p}{p}}^{\infty} f_{\Theta}(\theta)  f_{\Delta}(\delta) d\theta d\delta   \\
&=& \frac{1}{p(1-p)} \int_{-\infty}^{\infty} f_{\Theta}(\theta) f_{\Delta}\left(\theta+\ln\frac{1-p}{p}\right) d\theta
\end{eqnarray}
where differentiation under the $\beta$-integral used Leibniz's integration rule,
\begin{equation}
\frac{d}{dz} \left (\int_{a(z)}^{b(z)} f(x,z)\,dx \right )= \int_{a(z)}^{b(z)}\frac{\partial f}{\partial z} f(x,z)\,dx + f(b(z),z)\frac{db}{dz}-f(a(z),z)\frac{da}{dz}.
\end{equation}
\qed
}

Given the distribution of person and item parameters, an expected accuracy score can be thus derived. In Table~\ref{examples} we illustrate some simple examples of expected accuracy estimates based on normal distributions for both person and item parameters in a unidimensional model. We see that accuracy is very high for tests that are too easy or difficult, but for tests in which the items are well matched to the persons, the expected accuracy is substantially lower. The purpose of this illustration is to demonstrate that high prediction accuracy is not only a function of the estimation algorithm. Outcomes on a test where items are matched to person abilities will necessarily be harder to predict than in the case where the test is too easy or too hard for the population of test-takers. Moreover, prediction accuracy cannot, except by chance, exceed these theoretical upper bounds. 

\begin{table}
\centering
\caption{Examples of expected accuracy estimates.}
\begin{tabular}{ccc} \toprule
$\theta$ & $\beta$ & expected accuracy $S$\\ \midrule
\multirow{4}{*}{$\mathcal{N}(0, 1)$} & $\mathcal{N}(0, 1)$ & 0.7252\\
& $\mathcal{N}(0, 2)$ & 0.7946\\
& $\mathcal{N}(-5, 1)$ & 0.9833\\
& $\mathcal{N}(5, 1)$ & 0.9833\\ \bottomrule
\end{tabular}
\label{examples}
\end{table}





\section{Data applications}

We begin here with some small simulation studies that illustrate model selection and parameter recovery using the proposed methods and also provide some initial comparisons with the JML package in R \cite{chen2019joint, Chen2019a}. These simulation studies are intended to show that the proposed method works as intended in relatively simplified settings. We do not attempt a large-scale comparison among competing methods for MIRT. Following the simulations, we present two real data examples. The first is a standard application of MIRT to a multidimensional assessment of physics concepts. Although the method is primarily predictive, we show that the high dimensionality of the learned model is consistent with related work using a factor-analytic approach. The second case is an application to a large, sparse dataset collected from a massive open online course. Here, we apply a novel validation approach to the predictive model. Namely, we show that it may be used ``recommender-style'' to identify which homework items have content relevance to examination problems. This last analysis makes use of auxiliary data from an open-book exam in the same online course.

\subsection{Simulation study: multi-unidimensional test}

We first present a simulation study in which a three-dimensional compensatory IRT model (3d2PL) is used to generate responses from items that load onto only one dimension of discrimination: that is, a multi-unidimensional test. Abilities were drawn for 1000 simulated persons from a  multivariate normal distribution $\mathcal{N}(\boldsymbol\mu,\, \boldsymbol\Sigma)$ with zero means and covariances, $\boldsymbol\mu = \boldsymbol 0$ and  $\boldsymbol\Sigma = I_3$. 60 items were simulated with intercept parameters drawn uniformly from the interval $[-2.5,2.5]$ and $\ln\mathcal{N}(0,\,0.5)$ distributed discriminations for one random index (all other discriminations were zero). 

Using the simulated response matrix, M2PL models $\Omega$ of dimensions $\{1, 2, 3, 4, 6, 9\}$ were explored using the algorithms described above. The set of $L_2$ regularization hyper-parameters was $\Lambda_j = \{0.02, 0.04, 0.06, 0.08\}$ for each model $\omega_j \in \Omega$. Parameter optimization was done using Adam \cite{Kingma2014}, a variant of stochastic gradient descent with adaptive learning rate initialized at $\gamma = 0.1$. To save memory and speed up convergence, gradients were computed on 10 batches of the training set. Cross-validation used 5 folds, leaving out 12 items at a time for testing, and we selected the model that had the highest area under the ROC curve (AUC). For measuring performance, we report accuracy (ACC), AUC, Goodman and Kruskal's \citeyear{goodman1954measures} lambda ($\Lambda_{GK}$; a measure of proportional reduction of error), and root mean squared error (RMSE).

Results are shown in Tables~\ref{results-striated} and~\ref{results-bootstrap}. We note that the striated and the elementwise methods have similar predictive performance. In both cases, the unidimensional model performed substantially worse, the 2-d model slightly better, while  models of dimension 3 and higher were fairly close. This suggests that under-fitting the data (i.e., with too simple of a model) results in bigger performance decrements than over-fitting the data (i.e., with too complex of a model). Moreover, the analytical  upper bound on accuracy using the known (simulated) parameters was 0.7664. Thus, the multidimensional models performed nearly optimally. 

\begin{table}[ht]
\centering
\caption{Exploratory results on a simulated dataset using the striated method}

\begin{tabular}{rrrrrr}
\toprule
 dim &  $\lambda$ &    ACC &    AUC &  $\Lambda_{GK}$ &   RMSE \\
\midrule
   3 & 0.04 & \bf 0.7452 & \bf 0.8141 & \bf  0.4013 & \bf  0.4149 \\
   4 & 0.04 & 0.7422 & 0.8111 &  0.3948 & 0.4161 \\
   6 & 0.06 & 0.7413 & 0.8107 &  0.3918 & 0.4204 \\
   9 & 0.06 & 0.7416 & 0.8107 &  0.3924 & 0.4206 \\
   2 & 0.02 & 0.7343 & 0.7993 &  0.3755 & 0.4205 \\
   1 & 0.06 & 0.7177 & 0.7761 &  0.3388 & 0.4346 \\
\bottomrule
\end{tabular}\\
\label{results-striated}
\end{table}

\begin{table}[ht]
\caption{Exploratory results on a simulated dataset using the elementwise method}
\centering

\begin{tabular}{rrrrrr}
\toprule
 dim &  $\lambda$ &    ACC &    AUC &  $\Lambda_{GK}$ &   RMSE \\
\midrule
   3 &  0.04 & \bf 0.7433 & \bf 0.8116 &  \bf 0.4025 & \bf 0.4170 \\
   9 &  0.06 & 0.7400 & 0.8085 &  0.3951 & 0.4222 \\
   4 &  0.06 & 0.7394 & 0.8083 &  0.3936 & 0.4223 \\
   6 &  0.06 & 0.7390 & 0.8078 &  0.3930 & 0.4224 \\
   2 &  0.02 & 0.7323 & 0.7967 &  0.3793 & 0.4219 \\
   1 &  0.04 & 0.7183 & 0.7776 &  0.3472 & 0.4330 \\
\bottomrule
\end{tabular}
\label{results-bootstrap}
\end{table}

The striated method took 46 seconds per model on a computer with a 2.6 GHz Intel Core i7 processor from 2016, and the elementwise method took 47 seconds.
As a comparison, for a smaller dataset with half as many persons (500) one third as many items (20), and $d = 6$, the mirt package \cite{chalmers2012mirt} took 100 seconds, while ours took 2 seconds. The mirt package implements a Metropolis-Hastings-Robbins-Monro algorithm \cite{Cai2010} that does not assume any prior over the item parameters and uses MCMC to sample person parameters and optimize the item parameters. Our implementation, available on GitHub\footnote{\url{https://anonymous.4open.science/r/913f850f-e12c-45b0-a897-41782504fb59/}}, uses TensorFlow.

\subsubsection{Parameter Recovery}
\begin{figure}
  \centering
  \includegraphics{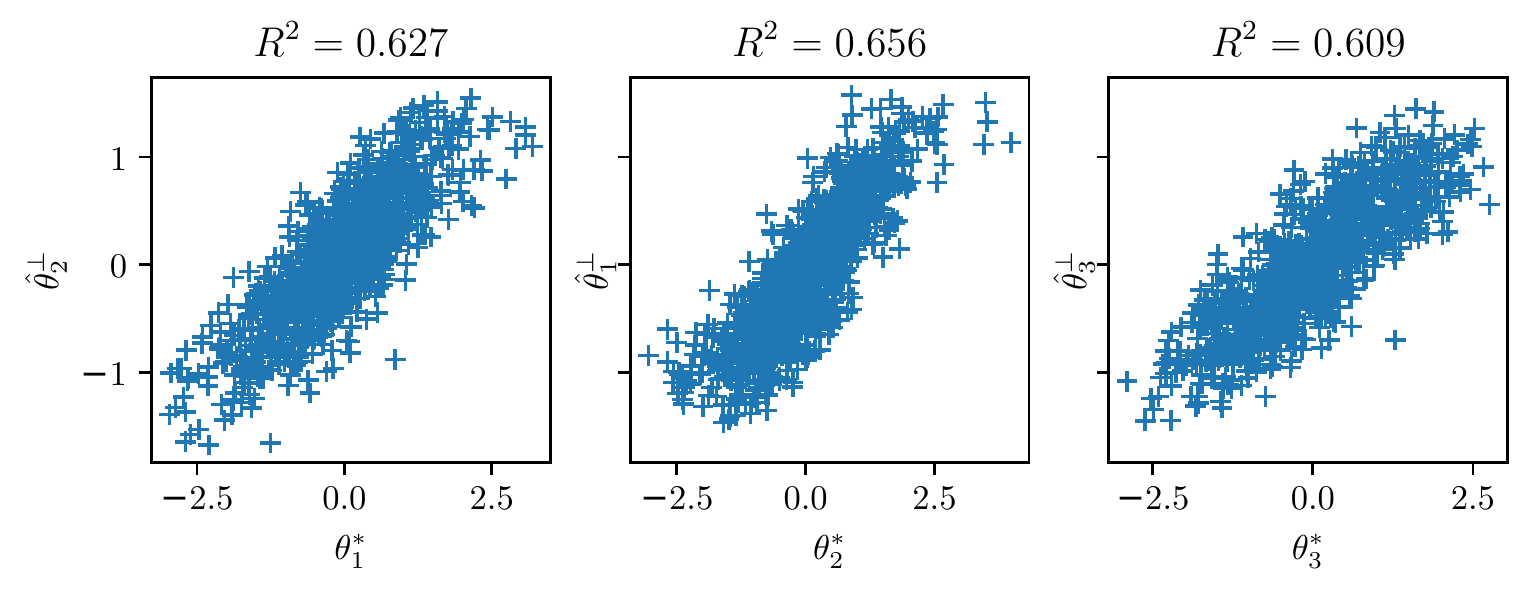}
  \caption{Proportion of variance explained by the estimated parameters, e.g. $\hat{\theta}^\perp_1$ is positively correlated with $\theta^*_2$.}
  \label{corr-plots}
\end{figure}

The exploratory search correctly identified the dimensionality of the model from which data were drawn ($d = 3$). We now consider whether the parameters themselves are recovered. Identifiability is still an issue as even with $L_2$ regularization terms, as a rotational invariance remains in the model. 
The estimated abilities may be orthogonalized by a standard procedure, using the projection matrix from the singular-value decomposition of the covariance matrix. Carrying this out, the now orthogonal ability vectors appear to correspond clearly to the simulated parameters up to an overall sign, see Table~\ref{corr-sim-orth}. The correlations between individual pairs of estimated and true values range from 0.83 to 0.84 in absolute value. Proportion of variance and the corresponding coefficients of determination are shown in Figure~\ref{corr-plots}.

\begin{table}[ht]
\centering
\caption{Correlation of estimated abilities with simulated values after orthogonalization}
\begin{tabular}{lrrrrrr}
\toprule
{} &  $\hat{\theta}^\perp_1$ &  $\hat{\theta}^\perp_2$ &  $\hat{\theta}^\perp_3$ &  $\theta^*_1 $ &  $\theta^*_2$ &  $\theta^*_3$ \\
\midrule
$\hat{\theta}^\perp_1$ &                    1.00 &                   -0.02 &                    0.00 &           0.26 &          \bf 0.84 &         -0.14 \\
$\hat{\theta}^\perp_2$ &                   -0.02 &                    1.00 &                    0.00 &          \bf -0.83 &          0.24 &          0.26 \\
$\hat{\theta}^\perp_3$ &                    0.00 &                    0.00 &                    1.00 &           0.23 &          0.10 &          \bf 0.83 \\
$\theta^*_1 $          &                    0.26 &                   -0.83 &                    0.23 &           1.00 &          0.02 &         -0.06 \\
$\theta^*_2$           &                    0.84 &                    0.24 &                    0.10 &           0.02 &          1.00 &          0.01 \\
$\theta^*_3$           &                   -0.14 &                    0.26 &                    0.83 &          -0.06 &          0.01 &          1.00 \\
\bottomrule
\end{tabular}

\label{corr-sim-orth}
\end{table}

In summary, in our simulation study using uncorrelated abilities and a multi-unidimensional test, our exploratory model was able to recover the correct model dimension in about 145 seconds of total run-time. After orthogonalization, ability estimates were strongly correlated with the simulated values. Figure 1 indicates that there was notable shrinkage of the latent trait estimates, which is to be anticipated in regularized settings and is also experienced with  IRT scoring methods that use a prior distribution for the latent traits \cite{Reckase2009}. 


\subsection{Simulation study: various levels of sparsity}

We ran benchmark comparisons against the method of constrained JMLE (CJMLE) in \cite{chen2019joint} with simulated datasets of various sizes and levels of sparsity. A 5-dimensional compensatory IRT model was used to generate responses from 80 persons over 20 items, or from 1000 persons over 50 items. Person abilities, item intercepts, and item discriminations were sampled from the standard normal distribution, and data was missing completely at random, with sparsity $p \in \{0, 0.2, 0.8\}$. In total, six such simulated datasets were generated. CJMLE and our own penalized JMLE method were compared in terms of likelihood, accuracy, area under the curve and run-time. The size of the estimated models was $d = 5$ for all datasets, except for the multi-unidimensional dataset described in the previous section, for which $d = 3$. Results for simulated datasets  are reported in Table~\ref{comparison_sim}. Our estimation code reaches higher log-likelihoods in all simulated cases. This observation is particularly significant when sparsity is high, as the hyper-parameters of CJMLE may restrict the parameter space too much when few samples are observed.

\begin{table}
  \centering
  \caption{Comparison with the constrained JMLE approach using simulated datasets of various size and sparsity}
  \scriptsize

\begin{tabular}{llllllllllll} \toprule
Data & \# persons & \# items	& Sparsity	& \multicolumn{3}{l}{Constrained JMLE}	& \multicolumn{3}{l}{Ours non regularized}			\\

& & & & \multicolumn{3}{l}{\cite{chen2019joint}} & \multicolumn{3}{l}{2 batches} \\

& &	&	& LL	& Acc	& AUC	& LL	& Acc	& AUC	\\ \midrule
Simulated 3-dim & 1000 & 60 & 0\% & -24085 & 0.79 & 0.87 &   -22530 & 0.81 & 0.89\\
\\
Simulated 5-dim & 80 & 20	& 0\% 	& -354	& 0.91	& 0.97		&   -244	& 0.95	& 0.98\\
& & & 20\% 	& -205	& 0.94	& 0.99		&  -76	& 0.98	& 1\\
& & & 80\% 	& -1.11	& 1	& 1	& \ -0.02	& 1	& 1\\
\\
Simulated 5-dim	& 1000 & 50 & 0\% 	& -18187	& 0.83	& 0.92		&   -17777	& 0.83	& 0.92\\
& &	& 20\% 	& -13189	& 0.85	& 0.93		&  -12880	& 0.85	& 0.93\\
& &	& 80\% 	& -1276	& 0.95	& 0.99		& -491	& 0.98	& 1\\ \bottomrule
\end{tabular}

  \label{comparison_sim}
\end{table}

\subsection{Correlated factors and cross-loading items}
The simulation may also be carried out using items with factor cross-loadings and correlated abilities. As an example, we have used a set of 30 three-dimensional items specified in a seminal text on multidimensional IRT \cite[Table 6.1]{Reckase2009} with parameters designed to be both realistic and challenging to estimate. Following the same reference, we simulate responses from 2000 persons, setting the mean vector for abilities to $[-0.4, -0.7, 0.1]$ and the variance-covariance matrix to 

\[
\begin{bmatrix}
1.21 & 0.297 & 1.232 \\
0.297 & 0.81 & 0.252 \\
1.232 & 0.252 & 1.96
\end{bmatrix}.
\]

Ability dimensions 1 and 3 are highly correlated in this case ($\rho_{13} = 0.8$). And the population has significantly lower mean ability along dimension 2.

\begin{table}[ht]
\centering
\caption{Exploratory results with correlated factors, striated method}

\begin{tabular}{rrrrrr}
\toprule
 dim &  $\lambda$ &    ACC &    AUC &  $\Lambda_{GK}$ &   RMSE \\
\midrule
   3 & 0.08 & 0.7015 & \bf 0.7611 &  0.2775 & 0.4441 \\
   6 & 0.08 & \bf 0.7038 & 0.7610 & \bf 0.2831 & 0.4439 \\
   9 & 0.08 & 0.7016 & 0.7608 &  0.2774 & 0.4440 \\
   2 & 0.06 & 0.7028 & 0.7595 &  0.2809 & \bf 0.4414 \\
   4 & 0.08 & 0.7005 & 0.7595 &  0.2752 & 0.4445 \\
   1 & 0.06 & 0.6974 & 0.7531 &  0.2670 & 0.4440 \\
\bottomrule
\end{tabular}
\\
\label{reckase-striated}
\end{table}

\begin{table}[ht]
\caption{Exploratory results with correlated factors,  elementwise method}
\centering

\begin{tabular}{rrrrrr}
\toprule
 dim &  $\lambda$ &    ACC &    AUC &  $\Lambda_{GK}$ &   RMSE \\
\midrule
   3 &       0.08 & \bf 0.7039 & \bf 0.7653 & \bf 0.3045 & \bf 0.4437 \\
   4 &       0.08 & 0.7003 & 0.7650 &  0.2962 & 0.4443 \\
   9 &       0.08 & 0.7014 & 0.7646 &  0.2986 & 0.4442 \\
   2 &       0.08 & 0.7019 & 0.7645 &  0.2998 & 0.4441 \\
   6 &       0.08 & 0.7007 & 0.7640 &  0.2969 & 0.4444 \\
   1 &       0.04 & 0.6937 & 0.7532 &  0.2803 & 0.4438 \\
\bottomrule
\end{tabular}

\label{reckase-bootstrap}
\end{table}

Results are shown in Tables~\ref{reckase-striated}-\ref{reckase-bootstrap} for exploratory models of  dimension $\{1, 2, 3, 4, 6, 9\}$ where hyper-parameters were selected in $\Lambda_j = \{0.02, 0.04, 0.06, 0.08\}$ during cross-validation.
With highly correlated factors and cross-loading items, the true 3-d model may or may not be selected as the best fit; the signal is not as clear. Using striated sampling, the 3-d model has the best AUC score, but the 2-d embedding has lower RMSE, and the 6-d model has higher accuracy and a higher Goodman-Kruskal lambda. Interestingly, the elementwise sampling algorithm selects a 3-d embedding as best on all quality-of-fit indices. \citeA{Reckase2009} chose these generating parameters in order to challenge the estimation software available at the time. In fact, it was found that even commercial MCMC-based estimation of a 3-d compensatory model ``gave what is essentially a two-dimensional solution for the data set'' (p. 175). We will consider these results again briefly along with findings from real data sets, which we turn to next.

\subsection{Physics concept test}

Our first real dataset used responses from $N \approx 13000$ students answering 30 items in conceptual physics on the Force Concept Inventory \cite{Hestenes1992}. There were very few missing entries, less than 0.2\%. We included models of dimension $\{1, 2, 3, 5, 7, 9, 20\}$. Running time took 111 seconds per model, including 22 seconds once the best hyper-parameter has been learned. Results are reported in Tables~\ref{results-striated-real} and~\ref{results-bootstrap-real}.

\begin{table}[ht]
\centering
\caption{Exploratory results on the FCI dataset using the striated method}
\begin{tabular}{rrrrrr}
\toprule
 dim $d$ &  best reg. $\lambda$ &     ACC &     AUC &  $\Lambda_{GK}$ &    RMSE \\ \midrule
\bf  20 & \bf      0.05 & \bf 0.7409 & \bf 0.8091 & \bf         0.4022 & \bf 0.4191 \\
\bf   9 & \bf      0.05 & \bf 0.7399 & \bf 0.8087 &  \bf        0.4000 &  \bf 0.4192 \\
   5 &       0.05 &  0.7402 &  0.8079 &          0.4005 &  0.4199 \\
   7 &       0.05 &  0.7384 &  0.8069 &          0.3964 &  0.4203 \\
   3 &       0.05 &  0.7367 &  0.8053 &          0.3925 &  0.4215 \\
   2 &       0.02 &  0.7343 &  0.8026 &          0.3872 &  0.4216 \\
   1 &       0.02 &  0.7250 &  0.7934 &          0.3654 &  0.4266 \\
\bottomrule
\end{tabular}

\label{results-striated-real}
\end{table}

\begin{table}[ht]
\centering
\caption{Exploratory results on the FCI dataset using the elementwise method}
\begin{tabular}{rrrrrr}
\toprule
 dim $d$ &  best reg. $\lambda$ &     ACC &     AUC &  $\Lambda_{GK}$ &    RMSE \\ \midrule
 \bf  9 & \bf      0.05 & \bf 0.7422 & \bf 0.8132 & \bf         0.4349 & \bf 0.4177 \\
\bf  20 & \bf      0.05 & \bf 0.7422 & \bf 0.8127 & \bf         0.4347 & \bf 0.4180 \\
   7 &       0.05 &  0.7395 &  0.8115 &          0.4292 &  0.4187 \\
   5 &       0.05 &  0.7400 &  0.8110 &          0.4300 &  0.4190 \\
   3 &       0.05 &  0.7384 &  0.8092 &          0.4263 &  0.4203 \\
   2 &       0.05 &  0.7373 &  0.8078 &          0.4240 &  0.4218 \\
   1 &       0.01 &  0.7268 &  0.7965 &          0.4015 &  0.4252 \\
\bottomrule
\end{tabular}

\label{results-bootstrap-real}
\end{table}

For both methods, models with $d = 9$ and $d = 20$ have comparable top performance, with hyper-parameter $\lambda = 0.05$. Using 20 dimensions to model 30 items comes perilously close to a 1:1 ratio, and we would hardly suggest modeling each item individually. The $d=20$ model was included here for two reasons. First, it shows that high dimensionality is tractable using this method.  Second, and more importantly, it demonstrates that the regularization procedure is working as it should. That is, out-of-sample prediction results improve with model complexity to a point, but then they level off. The $d=20$ model, properly regularized, does not significantly overfit the data. A case could be made for stopping, based on negligible improvement, at nine or even five latent dimensions. Interestingly, recent results from other authors using traditional exploratory methods independently identified a nine-factor model for the FCI \cite{Stewart2018}.

\subsection{Large-scale sparse MOOC dataset}

The MOOC dataset contains responses from $N \approx 30000$ students answering 197 homework items, for a total of 2 million entries. However, the sparsity of this dataset is 64\%, as many MOOC registrants participate only peripherally and answer very few homework items \cite{Seaton2014}. Although students were given multiple chances to answer homework questions, results from ``eventually correct'' scoring have very little variance, and it is often more informative to score each item based on correctness on the first attempt \cite{Bergner2015}. This is the approach we have taken here. We estimated models of dimension $\{1, 5, 10, 15, 20\}$. Running time took 7.5 minutes per model, including 88 seconds once the best hyper-parameter has been learned. Results are shown in Tables~\ref{results-striated-mooc} and~\ref{results-bootstrap-mooc}.

\begin{table}[ht]
\centering
\caption{Exploratory results on the MOOC dataset using the striated method}
\begin{tabular}{rrrrrr}
\toprule
 dim $d$ &  best reg. $\lambda$ &     ACC &     AUC &  $\Lambda_{GK}$ &    RMSE \\ \midrule
\bf  20 & \bf      0.02 & \bf 0.7648 & \bf 0.8323 & \bf         0.3935 & \bf 0.4005 \\
  15 &       0.02 &  0.7620 &  0.8293 &          0.3864 &  0.4022 \\
  10 &       0.02 &  0.7572 &  0.8233 &          0.3744 &  0.4058 \\
   5 &       0.01 &  0.7481 &  0.8140 &          0.3514 &  0.4113 \\
   1 &       0.05 &  0.7290 &  0.7898 &          0.3020 &  0.4263 \\
\bottomrule
\end{tabular}

\label{results-striated-mooc}
\end{table}

\begin{table}[ht]
\centering
\caption{Exploratory results on the MOOC dataset using the elementwise method}
\begin{tabular}{rrrrrr}
\toprule
 dim $d$ &  best reg. $\lambda$ &     ACC &     AUC &  $\Lambda_{GK}$ &    RMSE \\ \midrule
\bf  20 & \bf      0.02 & \bf 0.7641 & \bf 0.8333 & \bf         0.3974 & \bf 0.4005 \\
  15 &       0.02 &  0.7619 &  0.8305 &          0.3916 &  0.4022 \\
  10 &       0.02 &  0.7563 &  0.8239 &          0.3774 &  0.4062 \\
   5 &       0.02 &  0.7478 &  0.8148 &          0.3555 &  0.4114 \\
   1 &       0.01 &  0.7260 &  0.7862 &          0.3003 &  0.4248 \\
\bottomrule
\end{tabular}

\label{results-bootstrap-mooc}
\end{table}

For this dataset, $d = 20$ with $\lambda = 0.02$ is the top performing model. This MOOC dataset included a wider variety of items ($n=197$) administered over a period of 14 weeks, and to a diverse collection of participants including students from many countries and institutions, as well as teachers, professionals, and enthusiasts. Our application indicates that a high-dimensional model may be needed to explain the variance in such a large-scale dataset.

\begin{figure}[ht]
\centering
\caption{Exploratory results on all datasets}
\includegraphics[width=\linewidth]{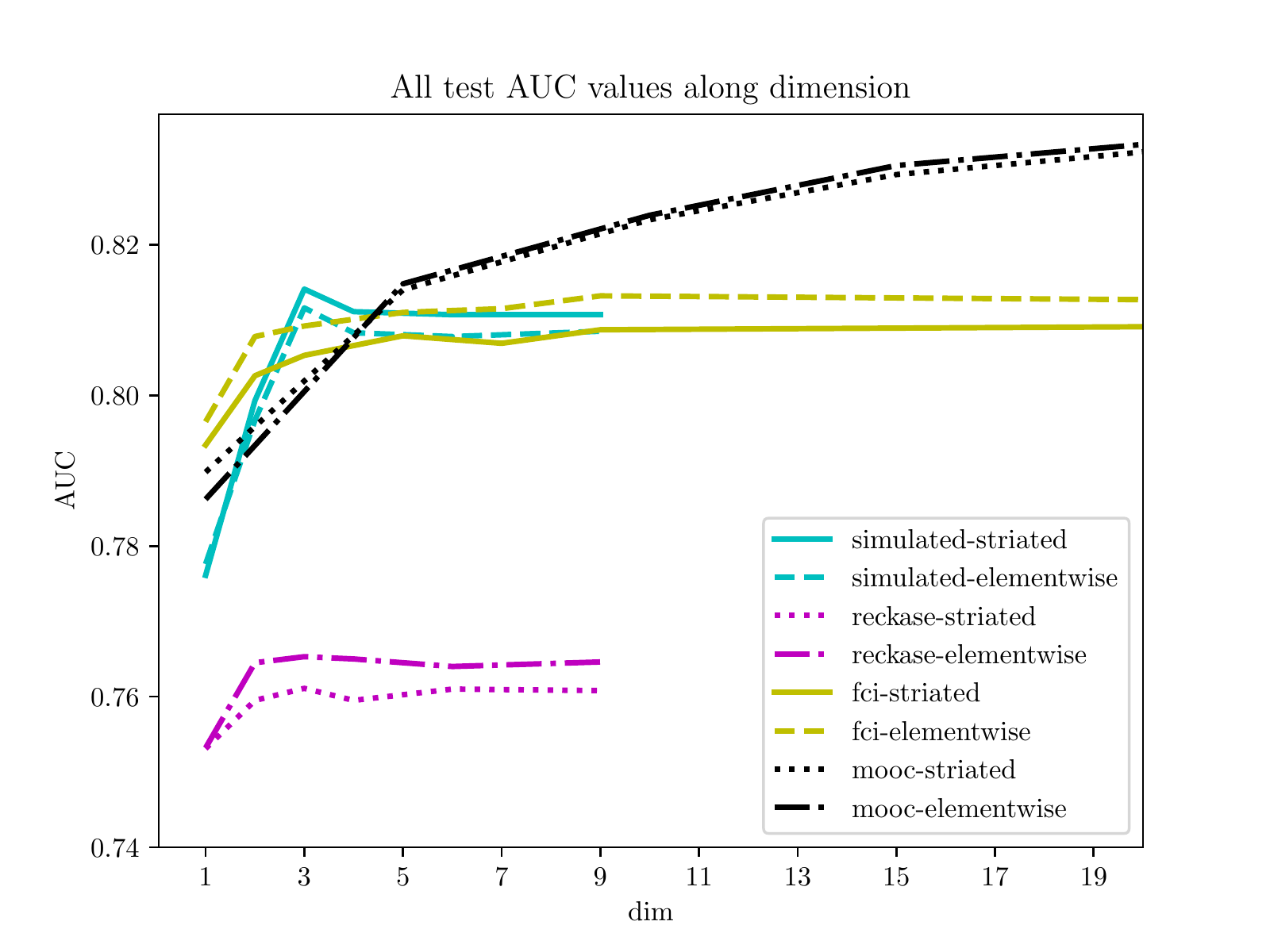}
\label{all-results}
\end{figure}

We have graphed the improvement in the AUC statistic from the exploratory results for all datasets in Figure~\ref{all-results}. For the simulated multi-unidimensional dataset, the performance peaks sharply at the true dimension $d = 3$. After this, performance steps down slightly and plateaus as dimension is increased. With responses simulated for correlated factors and cross-loading items (lines labeled `reckase'), we observe lower overall accuracy. Model AUC increases more gradually up to $d = 3$ and then plateaus, notwithstanding small fluctuations. The performance trend is quite similar for the Force Concept Inventory, with AUC flattening out and plateauing above $d=9$. Results for the MOOC dataset appear to be increasing at $d=20$ but with an indication of diminishing returns.

\subsubsection{Performance and run-time}

As for the simulated datasets, we include benchmark comparisons with constrained JMLE (CJMLE) for real datasets in Table~\ref{comparison_real}. For all models considered, $d = 5$.

\begin{table}
  \centering
  \caption{Comparison with the constrained JMLE approach using real datasets}
  \scriptsize

\begin{tabular}{lllllllllllll} \toprule
Data	& \multicolumn{4}{l}{Constrained JMLE}	& \multicolumn{4}{l}{Ours non regularized}				& \multicolumn{4}{l}{Ours regularized}\\
& \multicolumn{4}{l}{\cite{chen2019joint}} & \multicolumn{4}{l}{2 batches} & \multicolumn{4}{l}{$\lambda = 0.02$, 10 batches}\\
	& LL	& Acc	& AUC	& Time	& LL	& Acc	& AUC	& Time	& LL	& Acc	& AUC	& Time\\ \midrule
														
FCI Train & -116369	& 0.84	& 0.93	& $57''$	& \bf -113792	& 0.84	& 0.93	& $42''$	& -140406	& 0.82 & 0.91	& $12''$\\
FCI Test	& -31181	& 0.74	& 0.76	& 	& -49723	& 0.73	& 0.75	& 	& \bf -21461	& 0.75	& 0.77\\
\\																
MOOC Train	& -773746	& 0.79	& 0.88	& $33'18''$	& \bf -771703	& 0.79	& 0.88	& $5'12''$ & -843627 & 0.79 & 0.86 & $2'58''$\\
MOOC Test	& -108280	& 0.74	& 0.79	& 	& -109805	& 0.74	& 0.8 & & \bf -104761 & 0.74 & 0.80\\ \bottomrule
\end{tabular}

  \label{comparison_real}
\end{table}

We find that regularization results in a better reconstruction of missing data than benchmarked approaches: the unregularized model reaches higher likelihood on the training set than CJMLE but not on the test set---in other words, it overfits. The regularized model reaches higher likelihood than CJMLE on the test set, as the hyper-parameter $\lambda$ was selected using cross-validation. The speed up in performance compared to CJMLE is particularly noticeable on the high dimensional MOOC dataset, where our non-regularized approach converges to a higher likelihood in $\sim$5 minutes compared with $\sim$30 minutes for the R/C++ implementation of \cite{chen2019joint} available in the mirtjml package. The proposed algorithm is fast, partly thanks to early stopping, but also because mirtjml keeps the whole matrix of errors in response patterns in memory before applying the mask of missing data, while our implementation benefits from sparse computations. Our use of batching provides a trade-off between memory used at each iteration and number of iterations before stopping. That said, it is possible that improvements in convergence time and cross-validation error may be achievable by further adjusting the tolerance and constraining parameters in the CJMLE implementation.

\subsection{Validation using auxiliary data}

In the standard approach to exploratory factor analysis, one examines the items with similar loadings in order to identify or define the factors themselves. It is common for the analyst to have some expectations in advance about this structure, which is typically low-dimensional. In the case of the MOOC dataset, however, a 20-dimensional latent space solution was shown to have improved matrix-completion performance compared to smaller models. What does one make of this high-dimensional representation? Examining a loadings matrix with 20 columns and hundreds of rows would be extremely challenging and laborious. In this section we propose an alternative and automatable way to show the utility of the high-dimensional model in educational contexts even if the individual factors themselves are not clearly interpreted. Our approach is inspired by recommender-system applications of collaborative filtering, in contrast to the factor analytic approach. And in the case of the MOOC data, it makes use of auxiliary data---model-free observations---to ``validate'' the substantive content relationships of items with similar factor loadings.

Specifically, using web navigation events that occurred during the MOOC's open-book midterm examination, we determine the prior homework items that were most frequently consulted. We take the students' reference to these items (with solutions present) to suggest that they were substantively related to the exam content. This set of highly popular items then forms a reference set. Separately, the high-dimensional model can be used to produce an a set of items which have high overlap with factors that were prominent on the midterm. Finally, these two sets are compared. If the factor-overlap method recovers many of the same items students actually consulted (i.e., popular items), we can imagine that this information could be harnessed in a recommender system. That is, these homework items could have been recommended to students, for example in preparing for the exam. We now turn to the details of the auxiliary data and identification of the comparison item sets. 

Identification of the popular reference items required parsing the web navigation logs for each student. 
The MOOC midterm consisted of five problems with 26 separately scored items. Students had 24 hours in which to complete the exam, to account for international time-zone variation. In this exam, students were allowed to consult any course resources, including lecture videos or textbook pages. The largest class of resources, by frequency of navigation events, were in fact homework problems from the first half of the course. A navigation event consists of an origin-page, a destination-page, and a timestamp. For each student, only events that took place during their midterm ``window'' (from first access to final submit) were counted. Taken all together, these navigation events may be thought of as a network of course resource nodes. An edge from resource A to resource B is directed and weighted by the number of times a student navigated from one to the other during the midterm. If the same student traverses this path multiple times, all of the times were added to the edge weight.

To reduce noise in the transition network, we restricted our analysis to the 7518 students who participated in both the midterm and the final exam. These students likely took the MOOC seriously, whereas tens of thousands of others used it only sporadically. Even so restricted, there were over half a million navigation events between 1400 unique resources, including course lecture videos and the online textbook. Filtering this network to (a) course assessment items and (b) only the highest-weighted edges resulted in a set of 17 items, out of 102 items from the first half of the course. This is the set of high-popularity items. We note that edge weights drop off rapidly, similarly to a Pareto distribution. Our procedure makes use of a frequency cutoff, and a lower cutoff would have retained more items would be the reference set. Our proportional cutoff (0.05\% of all transitions) was designed to select a manageable set, roughly the top quintile of items.

The high-dimensional, penalized MIRT model was estimated using the full course dataset, including midterm and final exam items in addition to homework items. It is important to note that, although the whole course was used to fit the multidimensional model, roughly half of the course assessment items were available only after the midterm. Since multiple midterm items were grouped on the same page---thus making navigating events non-distinguishable at the level of midterm items---we aggregated the midterm items together to identify the most relevant factors. That is,
the 26 midterm item loadings (i.e., discriminations) were first discretized at a threshold value of 1.3 and then summed up. The resulting item-counts for each factor ranged from 0 to 18, and we retained the eight factors with counts above a threshold of 5. These eight factors were used to filter items from the rest of the assessment item pool. For each factor, the same discrimination threshold was used to select items. In total, a set of 41 non-midterm items were thus identified. However, only 12 of these were from the 102 pre-midterm homework items. The remaining 29 were future to the midterm. Thus, among items that were available for reference to students taking the midterm, 12 items were found to have high overlap on at least one of the factors dominant in the midterm. 90 out of 102 items from the pre-midterm period were found by this method to have low factor overlap. (As before, a lower cutoff would have retained more factors and thus more items).

\begin{table}
\centering
\caption{MOOC pre-midterm assessment items cross tabulated. Popularity refers to frequency with which items were referenced during midterm exams. Overlap refers to threshold comparisons of factor loadings with the midterm taken as a whole.}
\label{tab:convergent}
\begin{tabular}{@{}lll@{}}
\toprule
 & high popularity & low popularity \\ \midrule
high overlap & 11 & 1 \\
low overlap & 6 & 84 \\ \bottomrule
\end{tabular}
\end{table}

A summary of the comparison of the two item sets is shown in Table~\ref{tab:convergent}. We observe that 11 of the 12 high-overlap items are matched in the reference set of 17 high-popularity items. If we think of these as true positives, from a recommendation perspective, then we would describe the remaining counts as one false positive, six false negatives, and 84 true negatives. The overlap and popularity criteria in Table~\ref{tab:convergent} are manifestly associated. Thus, the results do indicate that common factor loadings from the whole course response matrix help identify a similar set of items to the ones students use most as reference during an exam. The convergence of these two approaches, we suggest, makes the high-dimensional predictive model indeed ``useful.''

Our approach here is meant only to provide a plausibility check on the sense-making potential of the multidimensional factor results. We acknowledge that this method used some arbitrary cutoffs regarding weights and loadings to keep the comparison manageable. Certainly, robustness claims cannot be made without a sensitivity analysis with respect to cutoffs, but this is beyond our current scope.

\section{Summary and discussion}

We have presented a novel computational approach, inspired by collaborative filtering, for fitting predictive multidimensional item response models. We optimize the joint maximum likelihood, penalized by $L_2$ regularization terms to counterbalance the number of item and person parameters. We described two methods for cross-validated estimation, the striated and elementwise algorithms, that achieve comparable performance in similar run-time. We showed that this method can effectively and efficiently estimate parameters of highly multidimensional models, even when the observed response data are sparse. The algorithms performed favorably in benchmark comparisons, especially with large and/or sparse datasets. In the work presented, we have concentrated on point estimates. We note with an eye toward future work that, using variational inference, the method can be extended to the Bayesian setting.

Considering the evaluation of predictive, multidimensional models, we have provided results of two different kinds. First, we derived some new analytic findings on the expected performance of commonly used accuracy measures. In comparing predictive models of increasing dimensions, we reported common evaluation metrics, such as accuracy and RMSE, as well as Goodman and Kruskal's lambda, an appropriate measure of predictive association between categorical variables. Second, in the application to data from a massive open online course (MOOC), we showed that a 20-dimensional predictive model can be useful from a recommender-system perspective. We note that the factor structure was derived from correct-incorrect scoring of person-item interactions, while the reference set of popular items was based on model-free observations of information seeking and not informed by correctness at all. Therefore, the convergent results observed in the MOOC case point to a substantive commonality in items identified by factor loadings of the predictive model.


The occurrence of 20-dimensional factor models is unusual in the psychometric literature but not necessarily so in machine learning approaches to collaborative filtering. For educational psychologists, for example, the factor structure of response data is usually assumed or designed to be simple. In exploratory studies, interpretability is a requisite for maintaining a factor in the model. Collaborative filtering is typically employed for different ends, such as in recommender systems engineered for massive and sparse datasets. A number of computational advances have been made to address the complexity of estimation in those applications. We have suggested that some of these ``tricks'' can also benefit psychometric applications, especially with response data obtained from learning environments as opposed to standardized tests or scales. 





\vspace{\fill}\pagebreak

\bibliographystyle{apacite}
\bibliography{A-M,N-Z}


\end{document}